%% file: main.tex
\documentclass[conference]{IEEEtran}
\usepackage{times}
\usepackage[numbers]{natbib}
\usepackage[bookmarks=true]{hyperref}
\usepackage{multicol}

\usepackage{graphicx}
\usepackage{amsmath}
\usepackage{amsthm}
\usepackage{amsfonts}
\usepackage{multirow}
\usepackage{algorithm}
\usepackage{algpseudocode}
\usepackage{algorithmicx}
\usepackage{bbold}
\usepackage{caption}
\usepackage{mathtools}
\usepackage{booktabs}
\usepackage{hyperref}
\usepackage{float}

\input{math_commands.tex}
\usepackage{xspace}
\newcommand{\anameDipac}{DiPac\xspace}

\renewcommand{\eqref}[1]{(\ref{#1})}

\newcommand{\ie}{\textrm{i.e.}}
\newcommand{\eg}{\textrm{e.g.}}
\newcommand{\etc}{\textrm{etc.}}

\newenvironment{dm}{\vspace*{-2pt}\displaymath}{\vspace*{-2pt}\enddisplaymath}

\theoremstyle{definition}  

\newboolean{bcmt}
\setboolean{bcmt}{true}
\usepackage{ifthen}

\usepackage{color}
\definecolor{darkgreen}{rgb}{0,0.5,0}
\definecolor{fullred}{rgb}{0.95,.0,.1}
\definecolor{brown}{rgb}{0.65,0.16,0.16}
\definecolor{orange}{rgb}{1,0.5,0}

\newcounter{ccmt}

\newcommand{\comment}[3]{\ifthenelse{\boolean{bcmt}}{\addtocounter{ccmt}{1}
  \marginpar{\tiny\noindent{\raggedright{{\colorbox{#3}{\sffamily\textcolor{white}{#1
            [\arabic{ccmt}]}}}}} \color{#3}{#2} \par}}{}}

\begin{document}

\title{Differentiable Particles for General-Purpose Deformable Object Manipulation}


\author{Siwei Chen, Yiqing Xu, Cunjun Yu, Linfeng Li and David Hsu \\\\

National University of Singapore}

\twocolumn[{%
\renewcommand\twocolumn[1][]{#1}%
\maketitle

\begin{center}
    \centering

\begin{tabular}{c@{\hspace*{5pt}} c@{\hspace*{5pt}} c@{\hspace*{5pt}} c@{\hspace*{5pt}} c}
    \hspace{-8pt}
    \includegraphics[width=0.19\linewidth]{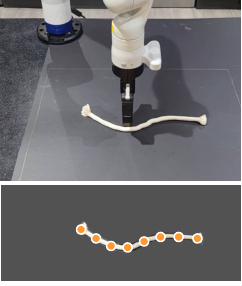} &
    \includegraphics[width=0.19\linewidth]{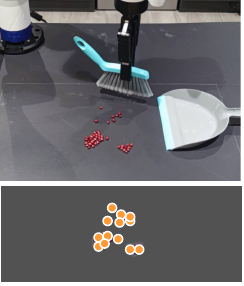} 
    &
    \includegraphics[width=0.19\linewidth]{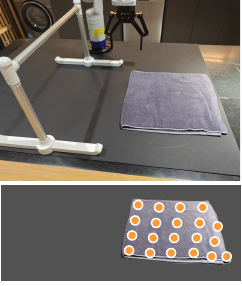} &
    \includegraphics[width=0.19\linewidth]{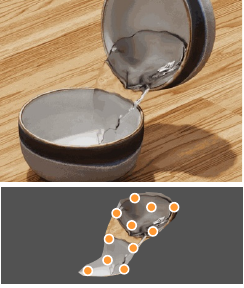} &
    \includegraphics[width=0.19\linewidth]{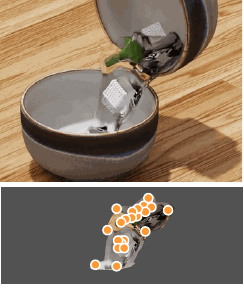} \\

    (\subfig a)  & (\subfig b)  & 
        (\subfig c)  & (\subfig d)  & (\subfig e) 
\end{tabular}

\captionof{figure}{Deformable object manipulation  tasks used in evaluation:
the first three  on a real robot and the last two in simulation.  
(\subfig a) Straighten the rope via pushing.  (\subfig b) Sweep the scattered
beans into a tray.
(\subfig c) Hang a piece of cloth on a rack.
(\subfig d) Pour liquid into a bowl.
(\subfig e) Pour soup, a liquid-solid mixture, into a bowl.
The top row shows the robot in action.
The bottom row shows the corresponding particle representations. }\label{fig:show_case}
\end{center}}]

\begin{abstract}

Deformable object manipulation  
is a long-standing challenge in robotics.
While existing approaches often focus narrowly on a specific type of
object, we seek a general-purpose algorithm, capable of manipulating
many different types of objects: beans, rope, cloth, liquid, \ldots.  One key
difficulty is a suitable representation, rich enough to capture 
object shape, dynamics for manipulation and yet simple enough to be acquired
effectively from sensor data.  Specifically, we propose \emph{Differentiable
  Particles} (\anameDipac), a new algorithm for deformable object
manipulation. \anameDipac represents a deformable object as a set of
\emph{particles} and uses a \emph{differentiable} particle dynamics simulator
to reason about robot manipulation. To find the best manipulation action,
DiPac combines learning, planning, and trajectory optimization through
differentiable trajectory tree optimization.  Differentiable dynamics provides
significant benefits and enable \anameDipac to (i)~estimate the dynamics
parameters efficiently, thereby narrowing the sim-to-real gap, and (ii) choose
the best action by backpropagating the gradient along sampled trajectories.
Both simulation and real-robot experiments show promising results.
\anameDipac handles a variety of object types. By combining planning and learning,
\anameDipac outperforms both pure model-based planning methods and pure data-driven learning
methods.
In addition, \anameDipac is robust and adapts to changes in dynamics, thereby
enabling the transfer of  an expert policy from one
object to another with different physical properties, \eg, from a rigid rod to a
deformable rope.

\end{abstract}

\IEEEpeerreviewmaketitle

\input{sections/introduction}

\input{sections/related_works}
\input{sections/method}

\input{sections/experiment}

\input{sections/discussion}

\input{sections/conclusion}

\bibliographystyle{plainnat}
\bibliography{references}

\clearpage
\appendix

\input{sections/appendix.tex}

\end{document}

%% file: math_commands.tex
\usepackage{amsmath,amsfonts,bm}
\usepackage{xspace}

\def\figref#1{Fig.~\ref{#1}}

\def\secref#1{section~\ref{#1}}

\def\eqref#1{equation~\ref{#1}}

\def\1{\bm{1}}

\DeclareMathAlphabet{\mathsfit}{\encodingdefault}{\sfdefault}{m}{sl}
\SetMathAlphabet{\mathsfit}{bold}{\encodingdefault}{\sfdefault}{bx}{n}

\DeclareMathOperator*{\argmin}{arg\,min}

\renewcommand{\eqref}[1]{(\ref{#1})}
\newcommand{\subfig}[1]{\textit{#1}}

\newcommand{\state}{\ensuremath{x}}
\newcommand{\initialState}{\ensuremath{x_0}\xspace}
\newcommand{\approxiState}{\ensuremath{\hat{x}}}
\newcommand{\goalState}{\ensuremath{x^\mathrm{g}}\xspace}
\newcommand{\action}{\ensuremath{u}\xspace}
\newcommand{\initialAction}{\ensuremath{u_0}\xspace}
\newcommand{\cost}{\ensuremath{c}}
\newcommand{\objectiveFunction}{\ensuremath{J}}
\newcommand{\chamferDistance}{\ensuremath{\sigma}\xspace}
\newcommand{\actionSeq}{\ensuremath{U}}
\newcommand{\horizon}{\ensuremath{H}}
\newcommand{\transition}{\ensuremath{f}}
\newcommand{\observation}{\ensuremath{o}}

\newcommand{\initialPolicy}{\ensuremath{\pi_\theta}\xspace}
\newcommand{\taskHorizon}{\ensuremath{T}\xspace}

%% file: sections/introduction.tex
\section{Introduction}

Deformable object manipulation (DOM) is a long-standing challenge in robotics,
because it involves a large degree of freedom and complex non-linear
dynamics.  Existing approaches often focus narrowly on a specific type of
objects: rope~\cite{sundaresan2020learning, lui2013tangled, nair2017combining},
cloth~\cite{ganapathi2021learning, cusumano2011bringing, lin2021learning, monso2012pomdp}, liquid~\cite{spnets2018, xian2023fluidlab, lee2023aquarium}, \ldots\,. Such a
strategy is unlikely to exhaust the wide variety of different deformable
objects.  It is essential to develop a \emph{general-purpose}  DOM algorithm that handles many
different types of objects uniformly.  Our work is a step in this direction.

One key difficulty is a suitable representation, which is  rich enough to capture the complex
object shape, object dynamics for manipulation and yet simple enough to be acquired
effectively from sensor data.
We  represent a deformable
object as a set of \emph{particles}, \ie, 3-D points with associated physical
properties, and use a differentiable particle dynamics
simulator to approximate object dyanmics.

The particle representation provides multiple benefits. First, it is general
and capable of representing a wide variety of objects: 0-dimensional
beans, 1-dimensional rope, 2-dimensional cloth, and 3-dimensional liquid
(\figref{fig:show_case}). Further, differentiable particle dynamics enables
efficient gradient-based optimization.
Finally, particle dynamics is computationally efficient with 
parallel implementation on modern GPUs. 

Our new algorithm, \emph{Differentiable Particles} (\anameDipac), takes
RGBD images of a deformable object as input and outputs a sequence of robot
manipulation actions, until a specified goal is reached.  To find the best
action, \anameDipac combines three elements---learning,
planning, and trajectory optimization---through differentiable trajectory tree
optimization. It adopts an uncommon policy representation for manipulation: a
computational graph consisting of a trajectory tree parameterized by 
the differentiable particle-based dynamics simulation model.  The entire computational graph is end-to-end
differentiable for both model calibration and action selection.
During the offline model calibration phase, the robot learns the dynamics parameters of the
simulator by interacting with the real object.

It is important to note that the key objective of our model calibration is to
enhance the model prediction accuracy for action selection rather than to
uncover the ground-truth physics.  During the online action selection  phase,
\anameDipac uses the calibrated dynamics simulator to reason about
manipulation actions. Guided by a learned initial policy, it searches the
trajectory tree for the best actions.  In both phases, differentiability plays
a critical role and allows \anameDipac to update the dynamics parameters and
find the best actions through backward gradient propagation.

The policy representation of \anameDipac is aligned in spirit with the
\emph{Differentiable Algorithm Network} (DAN) framework~\cite{karkus2019differentiable}. The gist of DAN is to
represent a robot control policy as a model and an associated algorithm that
solves the model, both encoded together in a neural network (NN). The NN
policy representation is fully differentiable and can be trained end-to-end
from data.  \anameDipac uses a particle dynamics simulator as the model and
a trajectory tree optimizer as the algorithm. While it does not use a NN for
policy representation, the trajectory tree plays the same role and is fully
differentiable.

We evaluated \anameDipac on tasks on several distinct DOM tasks:
straightening a rope via pushing, sweeping scattered beans into a
target region, hanging a towel on a rack, and pouring liquid or soup.
While specialized methods may perform better on any particular one of these
tasks, \anameDipac performs capably over all of them.
Furthermore, by combining planning and learning, \anameDipac outperforms
both pure model-based planning methods and pure data-driven learning
methods, including  two state-of-the-art
methods, Diffusion Policy~\cite{chi2023diffusionpolicy} and Transporter~\cite{zeng2020transporter}.
Finally, an ablation study shows that each constituent element of \anameDipac---learning, planning, and
trajectory optimization---is essential for good performance.

One main limitation of our current work is particle state acquisition.  We
assume that a set of representative particles can be acquired from RGBD images
of the object. This assumption may fail with complex objects or environments,
\eg, cloth, because of occlusion or self-occlusion.  Nevertheless, our
evaluation tasks (\figref{fig:show_case}) represent a substantial range of interesting objects
that \anameDipac succeeds on.

%% file: sections/related_works.tex
\section{Related Works}
\label{sec:related}

\subsection{Model-based DOM}
Model-based DOM methods have been studied widely ~\cite{cusumano2011bringing,
  monso2012pomdp,zaidi2017model}. They leverage a model that connects robot
action with object deformation, allowing the robot to anticipate the object's
responses and choose the right action accordingly.  The model may be manually
designed~\cite{cusumano2011bringing, zaidi2017model} or learned from
data~\cite{Hu20183d, Hu20183GPR, lin2022learning,yan2021learning,
  wang2019learning}.  It may also adapt to real-world dynamics
online~\cite{chi2022irp, Wang_RSS_23}.

\anameDipac connects with the model-based approach. It uses a general-purpose
particle dynamics simulator as the model, with the simulation model parameters
learned from data. One key advantage of this model is a single, unified
representation for deformable objects of many different types, compared with
existing methods that use object-specific representations.
\anameDipac shares this underlying idea with earlier
work~\cite{li2018learning}, but combines planning, trajectory optimization,
and learning to achieve much stronger results on a variety of deformable
objects with a real robot.   

\subsection{Data-driven DOM} 
With the advances in machine learning, data-driven DOM methods are gaining
success increasingly.  Many of them learn robot actions directly from expert
demonstrations~\cite{lee2015learning, lui2013tangled, zeng2020transporter,
  seita2021learning, salhotra2022learning}.  Some learn a latent
representation aimed at an object's underlying physical properties,
enabling the robot to reason about object behaviors in different
scenarios~\cite{sundaresan2020learning, ganapathi2021learning}.

One highly successful example of data-driven DOM recently is Diffusion
Policy~\cite{chi2023diffusionpolicy}, which sets a new benchmark for a broad
the spectrum of manipulation tasks, including some DOM tasks.

Despite the success, data-driven methods often struggle with
generalization to unseen scenarios, because of data scarcity, \ie,  the
difficulty of acquiring sufficient real-world robot demonstration data. 
\anameDipac tackles the difficulty by combining the strengths of data-driven
and model-based approaches. The particle dynamics simulation model serves as a
general physics prior for learning DOM tasks.

\subsection{Differentiable Dynamics} 
Differentiable dynamics simulators, such as
PlasticineLab~\cite{huang2021plasticinelab}, DiSECt~\cite{heiden2021disect},
DiffSim~\cite{qiao2020scalable}, and DaxBench~\cite{Dax}, offer a powerful
tool for robot manipulation. They have two distinct uses.  First, they may serve as
a dynamics model in various simulation, control, and learning tasks~\cite{de2018e2e,
  spnets2018,Toussaint2018DifferentiablePA}.  Second, they provide  a
simulation environment for learning robot control
policies~\cite{chen2022imitation,freeman2021brax,mora2021pods,xu2022accelerated}. 
Our use in \anameDipac belongs to the former. 

In both cases, a significant challenge is to bridge the gap between the
simulated and real-world dynamics~\cite{chebotar2019closing}.  One general
idea is to learn model parameters from real-world data~\cite{li2018learning,
  mitrano2023focused,schenck2018spnets}.  \anameDipac follows this idea and
calibrates the particle dynamics model, using data from a small number of
interactions with the object in the real world.  Compared with earlier
work, \anameDipac has  a distinct objective in  model calibration:
instead of uncovering the underlying  physics, it aims to narrow
the distance between the predicted and actual states in the \emph{observation} space,
for improved action selection. 

%% file: sections/method.tex
\section{Problem Formulation}
\label{sec:problem_formulation}

\anameDipac takes RGBD images of a deformable object as input and outputs a sequence of robot manipulation actions in task space, until a specified goal is reached.  

We represent a deformable object as a set of \emph{particles}. Each particle is a point of the object,  with 3-D position,  velocity,  mass, as well as other attributes that characterize the interaction of this particle with other particles and the environment. The set of particles fully determines the \emph{state} of the object.  We measure the similarity between two states using the Chamfer distance for shape matching over the positional components of the particles. Let $P$ and $Q$ be two point sets in 3-D. The Chamfer distance  between $P$ and $Q$ is
\begin{equation*}
    \chamferDistance\left(P, Q\right)= \frac{1}{|P|} \sum_{p \in P} \, \min_{ q \in Q}\|p-q\|^2_2+ \frac{1}{|Q|} \sum_{q \in Q} \, \min_{p \in P}\|q-p\|^2_2 .
\end{equation*}

We assume a differentiable  simulator that captures the dynamics of the
particle system:
\begin{dm}
   \state_{t+1} = \transition(\state_t, \action_t; \varphi ), 
\end{dm}
where $\state_t$ and $\action_t$ are  the particle  state and the robot action at time
$t$, respectively.  The set $\varphi$ contains the dynamics parameters describing, \eg, 
object rigidity,   environment friction, \etc\,. The function $f$ is differentiable
with respect to $x_t, \action_t$ and $\varphi$. By default, we implement $f$
using a recent \textit{material point method} (MPM)~\cite{hu2018moving,
  sulsky1994particle}, but there are many  alternatives  (Section~\ref{sec:related}).
Details on MPM are available in  Appendix~\ref{appendix_sec:dynamics}.
\anameDipac is given an initial state \initialState and goal state \goalState. It aims to find a sequence of actions $(\action_0, \action_1, \ldots)$ that minimize the total cost $J$ over  maximum  \taskHorizon time steps: 
\begin{equation}
    \objectiveFunction(\initialState, \actionSeq_{0:\taskHorizon-1}) = \sum_{t=0}^{\taskHorizon-1}\cost(x_t, u_t).
\end{equation}

To reach the goal state \goalState as closely as possible, we define the
following cost function
\begin{equation}
    \cost(\state_t, \action) = \chamferDistance(\state_{t} , \goalState) - \chamferDistance(\state_{t-1} , \goalState) + \zeta(\state_{t} , u_t),
    \label{eq:reward_func}    
\end{equation}
where the similarity to the goal state is measured using the Chamfer distance \chamferDistance, and contact cost $\zeta$ is the L2 distance between the gripper and object.

Our premise in this work is that only a limited amount of real-world expert demonstration data, such as 10 episodes, is available from real robots. This constraint reflects the practical challenges of data collection, including the logistical difficulties, the time and effort required for gathering and annotating data, and the potential strain on robotic hardware. Given these challenges, our focus intensifies on creating learning algorithms that can leverage this sparse data effectively, underscoring the need for models that are data efficient.

\section{Deformable Object Manipulation with Differentiable Particles}

\subsection{Overview}

\anameDipac presents a novel approach for manipulating deformable objects by integrating learning, planning, and differentiable trajectory optimization into a cohesive framework. Our approach operates in two phases: the learning phase, where model parameters are estimated to predict a better behavior of deformable objects for action selection, and the testing phase, which employs a guided differentiable planner to select optimal actions.
Central to our approach is the unique combination of a tree-structured planning mechanism, guided by learned policies, with differentiable dynamics for action optimization along trajectories. This integration effectively addresses common challenges, such as poor initialization and local minima, by leveraging learned policy initialization and random policy branching.
\anameDipac introduces a new perspective on policy formation, combining planning, learning, and differentiable optimization in a way that enhances both the efficiency and applicability of deformable object manipulation.

\begin{figure}[t]
\centering

\includegraphics[width=0.9\linewidth]{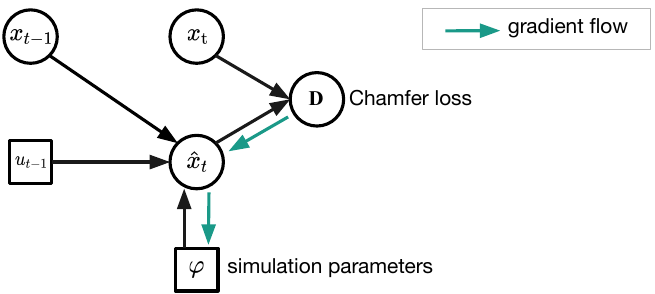}

\caption{
  Model calibration.
  The gradient-based optimization updates the parameter set $\varphi$ to minimize the Chamfer distance between the predicted state $\hat x_t$  and the observed particle state~$x_t$.
}
\label{fig:computation_graph}
\end{figure}

\subsection{Particle State Construction}
\label{sec:representation}

In this work, we assume there is no occlusion and reasonable object segmentation can be obtained from the top-down RGBD camera. 

\anameDipac acquires the particle states by reconstructing them in real-time from multi-view RGBD images. We first extract the particles from the depth image to create a point cloud, then use a segmentation mask from the RGB images to eliminate background points and retain only the object-related points. For simplicity and speed, we use color filtering but it can be substituted with a more advanced segmentation technique, for example, Segment Any Thing~\citep{kirillov2023segany, Qin_2020_PR}. The final point cloud of the object is obtained by merging points from each view, and the unobserved interior is then filled with uniformly sampled particles to model the deformable object. 
When a depth camera is not available, multi-view RGB images can still be used to reconstruct the object shape for manipulation tasks~\citep{ChenS-RSS-21}.

The particle-based state representation allows us to predict the future states using the material point method (MPM)~\cite{hu2018moving,sulsky1994particle}. In particular, the MPM simulation engine can roll out particle states to predict the next state $\state_{t+1} = \transition(\state_t, \action_t; \varphi)$, resolving the long-standing challenge of lack of accurate dynamics models for deformable objects, enabling model-based planning and model predictive control methods to be applied to more complex manipulation tasks.

\subsection{Dynamics Model Calibration} 
\label{sec:Dynamic_calibration}

To enhance the accuracy of the simulator, we update the dynamics parameters, the object's stiffness and the friction coefficient, using a small amount of interaction data from the real robot.  This calibration helps ensure that the actions optimized in the simulator produce similar results in the real world, essential for the successful deployment of \anameDipac in real-world applications. In detail, for each task, we collect 10 expert demonstrations from the real world via human teleoperation with random initial positions and poses of the deformable object. Those expert demonstrations consist of RGBD images and expert actions for each step. 

The differentiable particle representation enables direct optimization of the dynamics parameters through gradient descent. We illustrate this process in Fig.~\ref{fig:computation_graph}. A single trajectory consists of a sequence of observations $\observation_{0:\taskHorizon}$ and actions $\action_{0:\taskHorizon-1}$ from the real world, then is converted into a particle state trajectory $\state_{0:\taskHorizon}$. For each pair $(\state_t, \action_t)$ in $(\state_{0:\taskHorizon-1}, \action_{0:\taskHorizon-1})$, our model predicts the next state $(\approxiState_{1:\taskHorizon})$ in the simulation. To minimize the discrepancy between the simulation and the real world, we minimize the difference between the predicted states $(\approxiState_{1:\taskHorizon})$ and the real states $(\state_{1:\taskHorizon})$. As the particle-based transition function $\transition(\state_t, \action_t; \varphi)$ is differentiable, the dynamics parameters $\varphi$ of the simulator can be optimized through gradient descent, using the following objective function:

\begin{equation}
            \varphi = \argmin_{\varphi} \sum_{t=0}^{\taskHorizon-1} \chamferDistance( \transition(\state_t, \action_t; \varphi) , \state_{t+1}),
\end{equation}
where $\chamferDistance(.,.)$ is the Chamfer distance that measures the \emph{unordered set} distance between the predicted particles $\approxiState_{t+1}$ and the observed particles $\state_{t+1}$. In this way, we \emph{do not require the particle correspondence} for calculating the Chamfer distance. Minimizing the distance of all transition pairs allows us to align our simulated dynamics model with the observed deformable object behaviors. Our purpose is for better action selection instead of creating a simulation that achieves the ground truth dynamics model.

\subsection{  Integrating Learning, Planning, and Optimization}
\label{sec:diff_traj_opt}

\begin{figure}[t]
\centering

\includegraphics[width=0.9\linewidth]{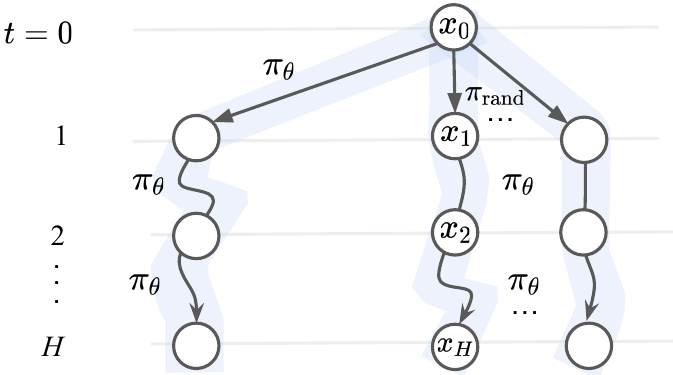}
\caption{
  Action selection via differentiable trajectory tree optimization.
  The search for the best action is guided by a learned initial policy
  $\pi_\theta$, together with random exploration policy $\pi_{\text{rand}}$.
  After rollouts, \anameDipac minimizes the total trajectory cost by back-propagating the gradient along each sampled trajectory.}
\label{fig:new_planner}
\end{figure}

With the learned dynamics model, we can search for the best actions. We introduce the Trajectory Tree Optimization algorithm to optimize trajectories sampled from a learned initial policy. This planner leverages differentiable trajectory optimization to achieve low sample complexity and efficient updates by pushing gradients directly. However, gradient-based methods can get stuck in local optima and are highly dependent on initial conditions~\cite{Dax}. To address these limitations, our planner uses a tree structure to select the best trajectory over multiple trajectories, initialized by randomly sampling a few initial actions (as depicted in Figure \ref{fig:new_planner}). This diversifies the initial states of the trajectories, helping to escape local optima and improve overall planning performance.

We implement our planner with a planning horizon $\horizon$. Note that this planning horizon $\horizon$ is smaller than the task horizon $\taskHorizon$ to save computation. At each time step $t=0,...,\taskHorizon-1$, our planner plans a locally optimal trajectory with horizon $\horizon$ from the current initial state and executes the first planned action. It then repeats this process and re-plans at the next state until it exhausts the task horizon $\taskHorizon$. 

We formalize the objective function for our trajectory tree optimization below. Given the current initial state $\initialState$ and a sequence of actions $\actionSeq_{0:\horizon-1} = (\initialAction, \action_1, ..., \action_{\horizon-1})$ of length $\horizon$, the total cost of the induced trajectory is the sum of the step-wise cost for each state-action pair. The objective function is to find the action sequence $\actionSeq^*_{0:\horizon-1}$ that minimizes this total cost:
\begin{equation}
    \actionSeq^*_{0:\horizon-1} = \underset{\actionSeq_{0:\horizon-1}}{\arg\min} \;\;\objectiveFunction(\initialState, \actionSeq_{0:\horizon-1}) =  \sum_{t = 0}^{\horizon-1} \cost(\state_t, \action_t),
    \label{equ:objective_original}
\end{equation}
whereas $\state_{t+1} = \transition(\state_t, \action_t; \varphi)$ is given by transition function.

Our planner inherently supports close-loop planning: for each time step, after executing the first control from the optimized control sequence $\actionSeq^*_{0:\horizon-1}$, it will start re-planning at the new state.

Unlike standard trajectory optimization algorithms, our planner takes advantage of a learned initial policy to make the trajectory optimization more efficient. In particular, we use a learned policy to initialize the sampling distribution of the trajectory, increasing computational efficiency and reducing sample complexity. Directly optimizing the objective function in equation (\ref{equ:objective_original}) is computationally intractable in the high-dimensional continuous state space. Even the naive Monte Carlo Tree Search in such a continuous action space requires a large amount of simulations for a good approximation. To make the optimization tractable, we use an initial policy to guide the trajectory rollouts and directly optimize the rollout trajectories through the gradients. A good policy can either initialize us with a near-optimal trajectory or enable us to optimize the gradients on a good optimization landscape. Either way, learning a good initial policy can help our planner to find the optimal solution more efficiently.

 \begin{figure*}[t]
\centering
\begin{tabular}{ccc}

    \includegraphics[width=0.31\linewidth]{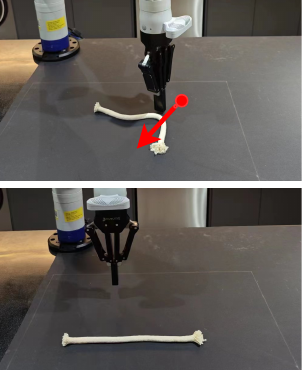} &
    \includegraphics[width=0.31\linewidth]{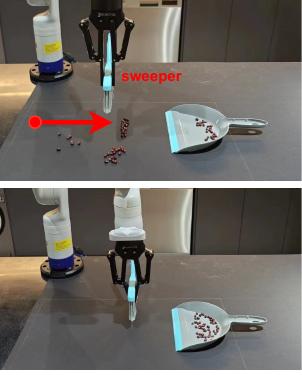} &
    \includegraphics[width=0.31\linewidth]{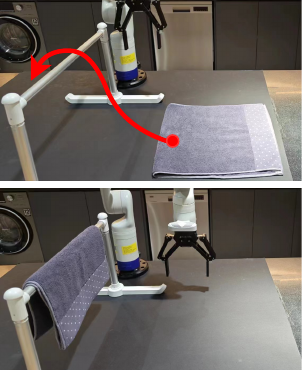} 

\end{tabular}
\caption{ Real-robot experiments setup. The top row shows robot actions. The bottom row shows the desired goal states.}\label{fig:robot_setup}
\end{figure*}

The trajectory tree, as shown in Figure~\ref{fig:new_planner}, balances sample complexity, computation speed, and final performance. We first generate $k$ random actions, $\initialAction^{1:k}$, by uniformly sampling a particle as the starting point and an angle to create a vector in a random direction to form the action. These actions are used for exploration. Additionally, we generate one action, $\initialAction^0$, from an initial policy $\initialPolicy$ for exploitation. Next, we use $\initialPolicy$ to roll out for total $H$ steps from all $\initialAction^{0:k}$.

We then use gradient descent to optimize the actions over each trajectory, where the objective function is the total cost of the induced trajectory under the transition function $\transition(x,u, \varphi)$. The optimization is performed as follows:

\begin{equation}
\action_t \leftarrow \action_t - \eta \bigtriangledown_{\action_t} \objectiveFunction(\state_t, \actionSeq_{t:\horizon-1}),
\end{equation}

Where $\eta$ is the learning rate. As the transition function and objective function are fully differentiable, we can update the actions by simply pushing the gradients. The total of $k+1$ action candidates are optimized simultaneously and can be fully parallelized. The action that minimizes the objective function will be selected for real robot execution.

The differentiable trajectory tree optimization uses an initial policy to guide the search and approximate the objective function. \anameDipac's initial policy uses Transporter \cite{zeng2020transporter}, where images are rendered from particles as input and actions as output. \anameDipac is not limited to a specific choice of the initial policy.  Alternatively, we can also use a Reinforcement Learning (RL) policy or a rule-based heuristic policy as the initial policy.

%% file: sections/experiment.tex
\section{Robot Experiments}
We evaluated \anameDipac on three DOM tasks on a Kinova Gen3 robot
(\figref{fig:show_case}\subfig{a}--\subfig{c}). We further evaluated two
liquid or liquid-solid pouring tasks in simulation
(\figref{fig:show_case}\subfig{d}--\subfig{e})., as it is difficult to carry
out repeated pouring tasks on a real robot due to hardware limitations.

\begin{figure*}[t]
\centering
\begin{tabular}{c c c}
\footnotesize{~~~~\textsf{Rope Pushing}}& \footnotesize{~~~~\textsf{Bean Sweeping}}  &
\footnotesize{~~~~\textsf{Cloth Hanging}}\\
  \hspace{-15pt}
    \includegraphics[width=0.32\linewidth]{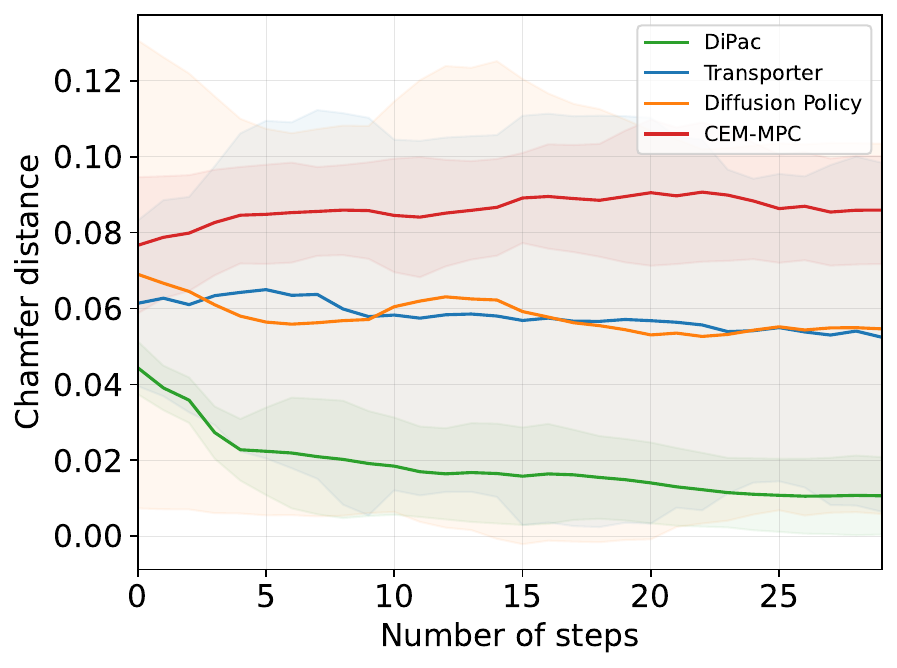} &
    \hspace{-10pt}
    \includegraphics[width=0.32\linewidth]{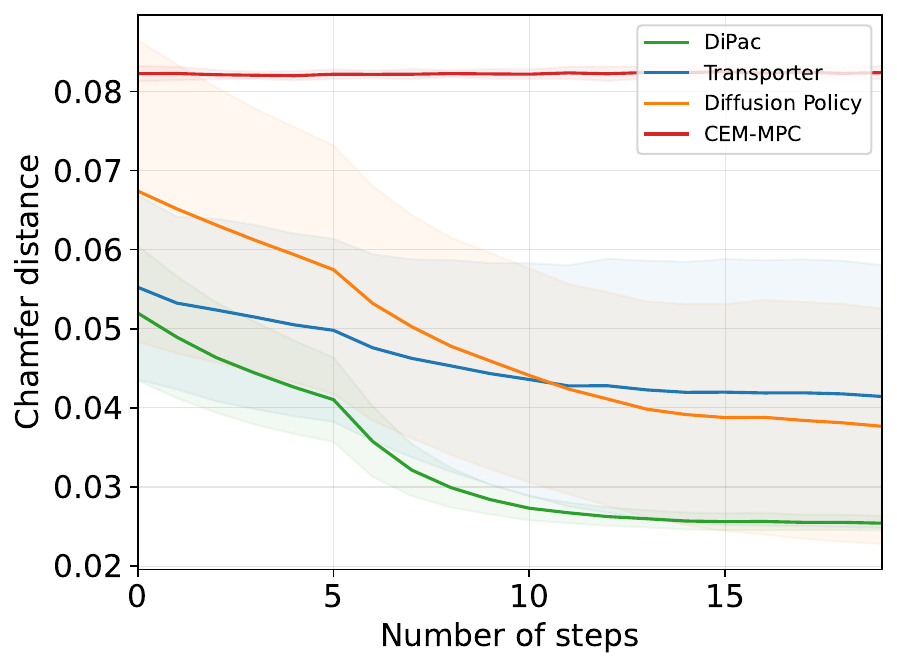} &
    \hspace{-10pt}
    \includegraphics[width=0.32\linewidth]{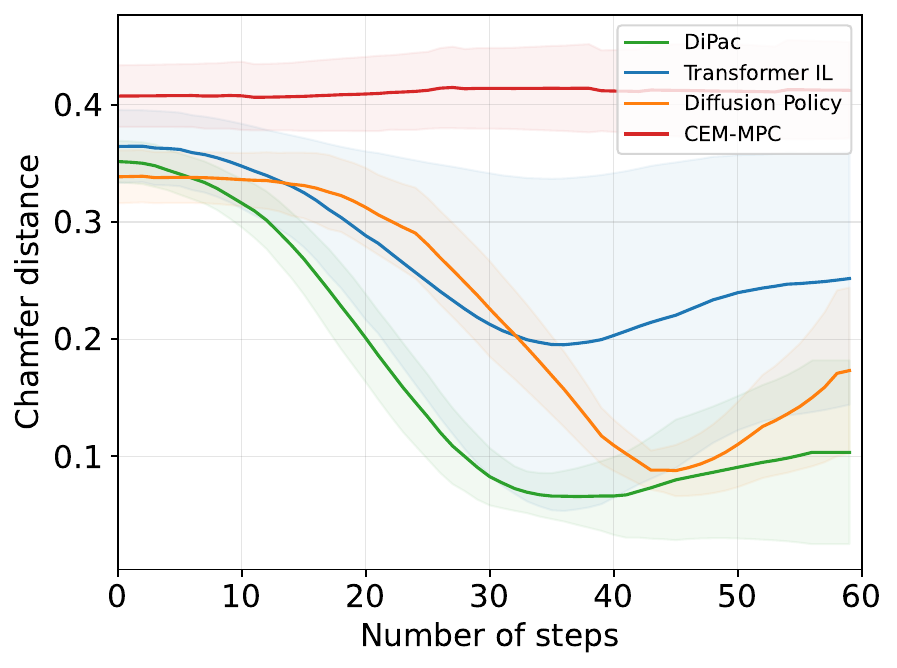} \\
\end{tabular}
\caption{ Real robot deformable object manipulation task results on three different materials. We report the averaged Chamfer distance (meters) over steps.  }

\label{fig:curve}
\end{figure*}

\begin{figure}[t]
\centering
\includegraphics[width=0.8\linewidth]{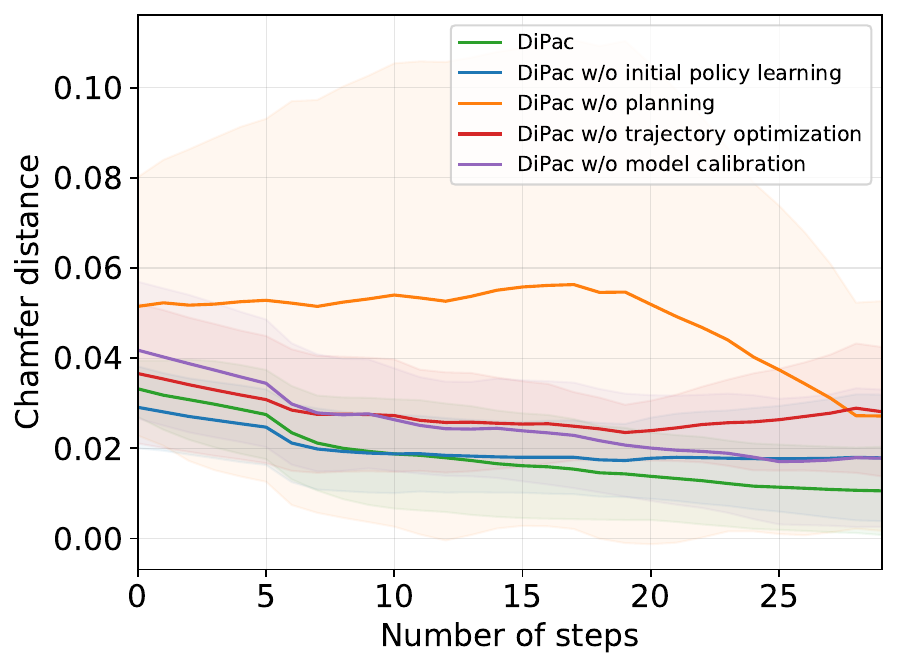}
\caption{ Ablation study on rope pushing. }
\label{fig:ablation_curve}
\end{figure}

We summarize the experimental findings below: 
\begin{itemize}
    \item \anameDipac performs well on the spectrum of different deformable
      objects under evaluation: beans, rope, cloth, liquid, and liquid-solid
      mixture (\figref{fig:curve} and Table~\ref{table:simulation}) . 
    \item \anameDipac outperforms a pure model-based planning method,
      CEM-MPC; it also outperforms two state-of-the-art data-driven learning
      methods, Diffusion Policy and Transporter  (\figref{fig:curve} and Table~\ref{table:simulation}) . 
    \item  Initial policy learning, planning, and trajectory optimization each
      contributes strongly to the best online action selection (\figref{fig:ablation_curve}).
\end{itemize}

We first describe the tasks and the task-specific implementation details. The state space for all tasks is covered in Section \ref{sec:problem_formulation}. For each task, the action space, and experiment setup are described below. Hyperparameters utilized by \anameDipac for each task are listed in Appendix Table \ref{table:real_robot_hyperparameters}. 
We collect 10 expert demonstrations from the real world via human teleoperation for each task with random initial positions and poses. Those expert demonstrations, which consist of RGBD images and expert actions, are used in the training of initial policy and all the data-driven baseline methods, as well as learning model parameters.

\subsection{Rope Pushing}
\label{sec:rope_push}

The task is to push the rope to reach 
a target configuration (\figref{fig:robot_setup}\subfig a).

\noindent \paragraph{Action Space}
Each action is specified by the starting
point for pushing, followed by a sequence of $20$ displacements,
i.e. $(x_0,y_0, \Delta x_0, \Delta y_0, ..., \Delta x_{19}, \Delta y_{19})$,
whereas $x_0,y_0$ specifies the start position of the gripper, and
$\Delta x_i, \Delta y_i$ is the displacement for each sub-pushing action. The
dimension of the action space is thus 42. Such a large action space enables
flexible maneuver skills, for example, curved push to match a target curve
shape.

\noindent \paragraph{Experiment Setup}
We repeat the test 10 times with random starting positions and
poses. The initial policy is obtained from an imitation learning method, Transporter~\cite{zeng2020transporter} and the task horizon is 30. The desired goal state is extracted from an RGBD image and provided to all the methods.

\subsection{Bean Sweeping}
The task is to sweep beans into a tray (\figref{fig:robot_setup}\subfig b).

\noindent \paragraph{Action Space} 
The action space is defined by the push-start point $(x_0,y_0)$ and the push-end point $(x_1,y_1)$ on the table. The dimension of the action space is $4$. 

\noindent \paragraph{Experiment Setup} The experiment setup is the same as the
rope pushing task,  except that the task horizon is 20. We use a fixed number of 50 red beans with random initial positions and configurations in the working space.

\subsection{Cloth Hanging}
\label{sec:cloth_hanging}

This task involves the dexterous manipulation of a piece of cloth in a 3D space, aiming to accurately hang it on a designated rack. This sophisticated task encompasses various challenges, including the accurate control of the robot gripper to grasp, transport, and release the cloth at precise locations and orientations to ensure that it remains on the rack.

\noindent \paragraph{Action Space}
The action space for this task is defined by a sequence of 8 small actions at each time step, with each action having a shape of $(8, 4)$. These actions represent the positions of the end-effector (robot gripper) in 3D space (3 dimensions for position and 1 dimension to signify the gripper's state, open or closed). The decision to open or close the gripper is crucial, particularly when releasing the cloth over the rack.

\noindent \paragraph{Experiment Setup}

The primary factor for success in this task is the stable placement of the cloth on the rack. A maximum task horizon of 60 steps is set to ensure efficiency in task completion while allowing sufficient flexibility for strategic maneuvering and adjustments. We assume the goal state is given and we evaluate 10 episodes for all methods with different starting positions and poses. 
A misjudgment of the release position can result in the cloth sliding off the rack and onto the table below, which results in increased distance to the goal state.  For this 3D manipulation task, Transformer-based IL \cite{zhao2023learning} has been used here for the initial policy.

\input{tables/simulation}

\subsection{Liquid and Soup Pouring}

 To further investigate the potential of DiPac, we conduct additional experiments that manipulate liquid in simulation. We use DaXBench~\cite{Dax} to simulate 2 robotic tasks: Pour Water and Pour Soup. 
 
 \noindent \paragraph{Action Space} The Pour Water task involves pouring water into a target bowl, using 6-D velocity and rotation controls in 100 steps. The Pour Soup task is similar but with solid ingredients, lasting 120 steps and using the same controls. The ground truth dynamics model is provided for all methods. These tasks present significant challenges as they require precise robot control of deformable objects, such as water, which are more difficult to manipulate due to their fluid and unpredictable nature. Achieving specific goals in a limited time frame is a major challenge in these tasks.

\paragraph{Baselines} We compare \anameDipac approach to Diffusion Policy, and baselines from the benchmark DaxBench~\cite{Dax} including ILD \citep{chen2022imitation}, a differentiable imitation learning method, as well as three variants of trajectory optimization methods that are classic non-differentiable Model Predictive Control and two differentiable implementations. Some data-driven methods, such as Transporter \cite{zeng2020transporter}, are limited to pick-and-place actions and are therefore not applicable to the fine-grained action tasks we examine here.

\subsection{Results}

The results for the tasks of Rope Pushing, Bean Sweep, and Cloth Hanging, as depicted in \figref{fig:curve}. It underscores the superior performance of \anameDipac over established baselines, including Transporter~\cite{zeng2020transporter}, a data-driven method for deformable object manipulation; Diffusion Policy~\cite{chi2023diffusionpolicy}, a state-of-the-art method for object manipulations including deformable objects; and CEM-MPC~\cite{Richards2005RobustCM}, a traditional non-differentiable trajectory optimization method.

\anameDipac significantly outperforms all baseline methods across the board. Its ability to efficiently handle complex, nonlinear dynamics and high degrees of freedom inherent in real robot deformable object manipulation tasks is clearly demonstrated. This is in stark contrast to the performance of the Diffusion Policy and Transporter methods, which exhibit considerable difficulties in adapting to unseen states during Bean Sweeping, Rope Pushing, and Cloth Hanging tasks. Although these imitation learning (IL) methods show competence in environments closely resembling their training demonstrations, their efficacy drops markedly in unfamiliar settings, leading to high variance in performance. This variance underscores the adaptability challenge that \anameDipac effectively overcomes, showcasing its robustness and the rapid convergence of its manipulation strategies.

In the task of Cloth Hanging, incorrect release positions result in the cloth slipping off onto the table, which is reflected in the end stages of the performance curve where the Chamfer distance tends to increase. This rise in the curve serves as an indicator of the manipulation policy's effectiveness; the lesser the increase, the more capable the policy at managing such precise tasks. \anameDipac's performance in this regard indicates its superior manipulation capabilities, further highlighting its advantage over the baselines.

The results for the tasks of Pour Water and Pour Soup are depicted in Table \ref{table:simulation}.  \anameDipac approach demonstrates superior performance compared to the baselines on the two liquid tasks. \anameDipac outperforms ILD by nearly double on the tasks of pouring water and soup. For the trajectory optimization methods, we improve performance consistently outperform these methods by a large margin on the two tasks. The manipulation of liquid to a specific configuration poses significant challenges in deformable object manipulation, resulting in high sample complexity and computation overhead for data-driven approaches and non-differentiable trajectory optimization. The results demonstrate that \anameDipac is a highly effective method for general DOM tasks, delivering promising performance compared to existing methods on these challenging DOM tasks.

In summary, \anameDipac not only surpasses its counterparts by a significant margin in all tasks but also demonstrates an exceptional ability to quickly and effectively converge to a good manipulation strategy. Its success in addressing the high variance seen in IL methods in unfamiliar scenarios and its proficiency in executing tasks requiring high precision,  emphasize its potential in real-world applications of deformable object manipulation.

\begin{figure*}[t]
\centering
\includegraphics[width=1.0\linewidth]{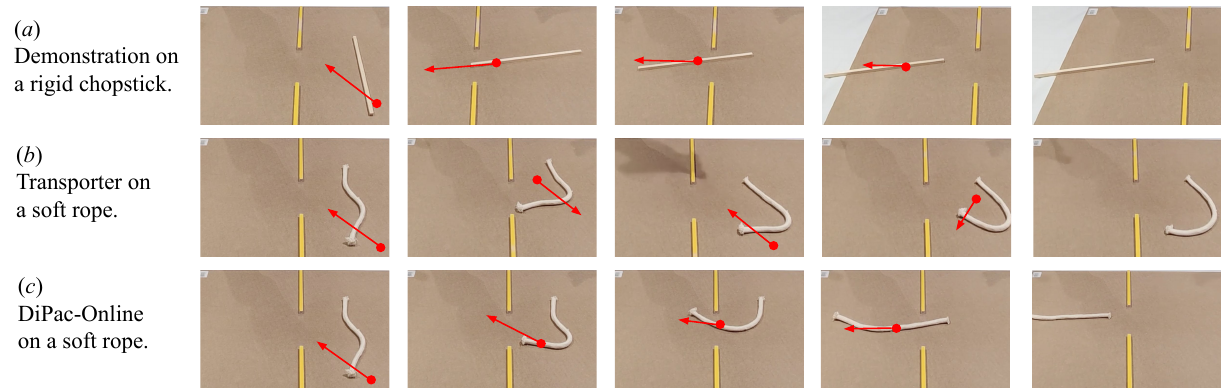}
\caption{
Policy transfer to a new domain. The task is to push an object through a narrow gap between obstacles.  (\subfig a) The expert demonstration pushes a rigid chopstick through by reorienting it. (\subfig b) Transporter tries to push through a soft rope but fails. (\subfig c) DiPac-Online succeeds, as it adapts to the changed dynamics by learning the dynamics parameters online.
}
\label{fig:transfer}
\end{figure*}

\subsection{Ablation Study}

To highlight the individual components' contributions of our combined planner with planning, learning, and differentiable optimization, we conducted an ablation study on the real robot manipulation task of pushing a rope. Especially, we aimed to elucidate the importance of i) incorporating a calibrated simulated dynamics model, ii) trajectory optimization based on differentiability, iii) exploration during the initial steps, and iv) having guidance from a learned IL policy. To do so, we incorporated these elements into four different systems and assessed them one by one.

\begin{itemize}
    \item DiPac w/o model calibration: We eliminate the differentiable model calibration procedure and substitute it with a random selection for the dynamics parameters (both rigidity and ground friction).
    \item DiPac w/o trajectory optimization: In this model, we exclude the differentiable trajectory optimization from our proposed planner, but keep the model calibration.
    \item DiPac w/o planning: We omit the random exploration at the initial steps, and conduct the optimization purely on the only trajectory roll out from the initial policy.
    \item DiPac w/o initial policy learning: We do not use any initial policy, rather a random policy is used to guide the proposed planner.
\end{itemize}

\noindent Overall, each component is critical for effective task performance. The results of the ablation study confirm the significance of every individual component that constitutes our proposed planner. Without differentiable optimization or exploration, the performance suffers remarkably, pointing at these two elements being critical for trajectory optimization in deformable object manipulation. Also, unless the dynamics gap is reduced, a drop in task performance is observed. This supports our claim of the necessity of each individual component, thereby establishing the novelty of our 'planning + learning + differentiable trajectory optimization' approach. This unique assembly not only introduces a novel perspective on forming policy but also leverages the benefits of each component to master complex manipulation tasks efficiently.

\subsection{Adaptation to New Dynamics }
Moreover, \anameDipac can easily adapt to new dynamics. We use \anameDipac-Online, an extension of \anameDipac to adapt to sudden dynamics changes through online calibration. \anameDipac-Online was initially trained on a rigid-rod pushing task and then transferred to a soft-rope pushing task, which has vastly differing dynamics. The video can be viewed in the supplementary files. Using the Transporter~\cite{zeng2020transporter} and \anameDipac methods, we set out to complete the task of pushing a rigid rod and soft rope through a small opening to the left area, shown in \figref{fig:transfer}. We start by collecting 10 expert demonstrations on the rigid rod to train both the Transporter and \anameDipac's initial policy. However, when the Transporter is transferred to roll out on the soft rope, it fails to complete the task due to the deformation of the rope which created states unseen in the training distribution. In contrast, \anameDipac is successful in completing the task, by continuously estimating the simulation parameters and calibrating the dynamics on the fly. \anameDipac can leverage the deformation of the rope and utilize planning to complete the task. This demonstrates the effectiveness of \anameDipac in tackling dynamics transfer scenarios.

%% file: tables/simulation.tex
\begin{table*}[t]
\centering
\begin{tabular}{lcccccc}
\toprule
 
   &
 
  \multicolumn{3}{c}{Planning} &
  \multicolumn{1}{c}{Ours} &
   \multicolumn{2}{c}{Imitation Learning} \\
\cmidrule(lr){2-4} \cmidrule(lr){5-5} \cmidrule(lr){6-7} 

  Task &

  CEM-MPC$^\dagger$ &
  Diff-MPC$^\dagger$&
  Diff-CEM-MPC$^\dagger$ &
  \anameDipac &
    ILD$^\dagger$ &
    Diffusion Policy 

   \\
\cmidrule(lr){1-7}

  Pour-Water &

  0.58$\pm$0.01 &
  0.30$\pm$0.00 &
  0.58$\pm$0.01 &
  \textbf{0.69 $\pm$ 0.01} &
    0.32$\pm$0.03 &
   0.66$\pm$0.17 
   \\
  Pour-Soup &
  0.56$\pm$0.01 &
  0.44$\pm$0.00 &
  0.55$\pm$0.02 &
  \textbf{0.76 $\pm$ 0.11} &
    0.42$\pm$0.12 &
   0.40$\pm$0.09 
   \\

\bottomrule
\end{tabular}
\caption{\small Simulation Results. Task performance is measured using the normalized reward. $\dagger$ We use numbers reported in the deformable object manipulation benchmark DaxBench~\cite{Dax}. More details of tasks can be found in Appendix \secref{append: simultion}. We report the mean and standard deviation evaluated with 20 seeds.}
\label{table:simulation}
\end{table*}

%% file: sections/discussion.tex
\section{Discussion}

\anameDipac has achieved impressive outcomes in manipulating a variety of deformable objects, such as rope, cloth, beans and liquid. This is possible due to its utilization of the particle representation, which is flexible and expressive enough to universally describe the underlying states of all deformable objects. Additionally, \anameDipac takes advantage of the recent improvement of differentiable dynamics simulators, enabling the sampled trajectories to be optimized through back-propagation, thus improving its computation speed and reducing the sample complexity of trajectory optimization.

Our current assumption is that there is no occlusion in the multi-view RGBD images. This allows us to directly extract the particle states from the segmentation masks of the deformable objects. However, this simplifying assumption could limit the performance of \anameDipac in more complex environments, where factors such as occlusions, motion blur, and background clutter make extracting the particle information challenging. Therefore, our next step is to develop a more robust and efficient algorithm for particle extraction, allowing us to apply \anameDipac to unstructured and more complex environments.

We would like to scale \anameDipac to tasks with more complex optimization landscapes in the future.  However, more challenging tasks with multiple local optima and non-smooth/discontinuous landscapes may require careful design of the gradient-based optimization algorithms to handle such scenarios. We conjecture that our initial policy guidance can be useful in cases, where a good initialization of the landscape can significantly reduce the computational cost of gradient-based optimization. This approach is analogous to AlphaGo~\cite{silver2016mastering}: both use a pre-learned policy or a value function to estimate the future returns of the current action, allowing for a shorter search depth without compromising much on the optimality. We intend to better understand the impact of initial policies and the optimization process on the overall system performance.

%% file: sections/conclusion.tex
\section{Conclusion}

In conclusion, we present \anameDipac, a general algorithm that enables manipulation of a wide range of deformable objects using differentiable dynamics in simulation and on real robots. \anameDipac represents deformable objects as differentiable particles, which helps to reduce the gap between simulation and reality. \anameDipac also provides a new perspective on how policy can be formed, by combining learning, planning, and differentiable trajectory optimization. The experiments demonstrate \anameDipac's ability to effectively perform diverse tasks, such as rope pushing, bean sweeping, and cloth hanging. Additionally, \anameDipac demonstrates exceptional transferability, addressing a current challenge in the understanding of deformable objects. Looking ahead, it will be crucial and valuable to explore the application of \anameDipac to more complex and diverse tasks, as well as to extend the method to handle objects with more intricate dynamics.

%% file: sections/appendix.tex
\subsection{Dynamics Transition Function}
\label{appendix_sec:dynamics}
The state-of-the-art deformable objects simulators such as PlasticineLab~\cite{huang2021plasticinelab} and DaXBench~\cite{Dax} utilize the MLS-MPM~\cite{hu2018moving} method. In a nutshell, the MLS-MPM consists of three main steps:

\noindent\textbf{Particle to Grid (P2G)}. In the process of P2G, the MLS-MPM first updates its deformation gradient $\mathbf{F}^{n+1}$ given the previous $\mathbf{F}^{n}$ and velocity gradient $\mathbf{C}^n$. It then computes the affine velocity $\mathbf{A}^n$ to update grid velocity $\mathbf{v}$ as shown below: 

\begin{align*}
    \mathbf{F}_p^{n+1}&=\left(\mathbf{I}+\Delta t \mathbf{C}_p^n\right) \mathbf{F}_p^n \\
    \mathbf{A}_p^n &= m_p \mathbf{C}_p^n-\frac{4 \Delta t}{\Delta x^2} \sum_p V_p^0 \mathbf{P}\left(\mathbf{F}_p^{n+1}\right)\left(\mathbf{F}_p^{n+1}\right)^T\\
    (m \mathbf{v})_i^{n+1}&=\sum_p w_{i p}\left\{m_p \mathbf{v}_p^n+ \mathbf{A}_p^n \left(\mathbf{x}_i-\mathbf{x}_p^n\right) \right \}.
\end{align*}

We denote $i$ as the grid index, $p$ as the particle index, and $n$ as the number of updates. The $\mathbf{P}$ involves object properties such as rigidity to affect the deformation process. MLS-MPM aggregates the velocities of all the particles into a grid, which enables modeling the interactions among all particles without the need to search for neighbors of each individual particle.

\noindent\textbf{Grid Operation (GO)}. The MLS-MPM then normalizes the grid velocity and computes boundary conditions (BC) as below:
\begin{align*}
    \hat{\mathbf{v}}_i^{n+1}&=(m \mathbf{v})_i^{n+1} / m_i^{n+1}\\
    \mathbf{v}_i^{n+1}&=\mathrm{BC}\left(\hat{\mathbf{v}}_i^{n+1}\right).
\end{align*}

\noindent\textbf{Grid to Particle (G2P)}. Finally, The MLS-MPM aggregates the velocity $\mathbf{v}$ from grids and converts it back to particles. It then updates the velocity gradient $\mathbf{C}$ and position $\mathbf{x}$:

\begin{align*}
 \mathbf{v}_p^{n+1}&=\sum_i w_{i p} \mathbf{v}_i^{n+1} \\
 \mathbf{C}_p^{n+1}&=\frac{4}{\Delta x^2} \sum_i w_{i p} \mathbf{v}_i^{n+1}\left(\mathbf{x}_i-\mathbf{x}_p^n\right)^T  \\
 \mathbf{x}_p^{n+1}&=\mathbf{x}_p^n+\Delta t \mathbf{v}_p^{n+1} .
\end{align*}
More details can be found in the MLS-MPM paper~\cite{hu2018moving}.

\subsection{Implementation Details}
\label{appendix_sec:visualization}
There are 5 tasks studied in this work. For Rope Pushing and Bean Sweeping, PlasticineLab~\cite{huang2021plasticinelab} is used as the differentiable simulation engine. For Cloth Hanging, Water Pouring and Soup Pouring, DaxBench~\cite{Dax} is used as the differentiable simulation engine. Both of them implement MLS-MPM as an underling physics engine.

\begin{figure}[ht]
\centering
\begin{tabular}{c c c}
    \includegraphics[width=0.29\linewidth]{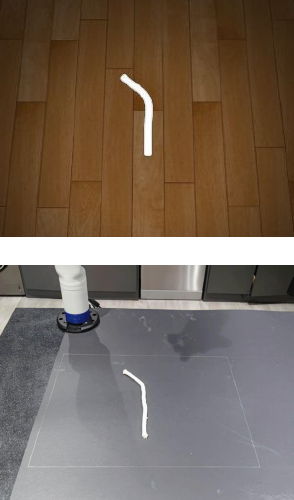} &
    \includegraphics[width=0.29\linewidth]{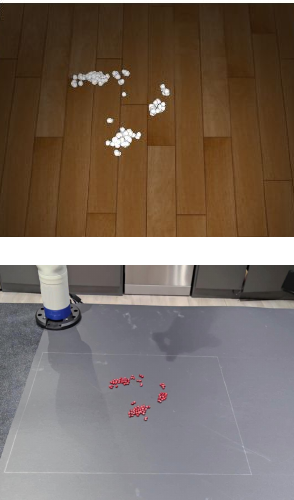} &
    \includegraphics[width=0.29\linewidth]{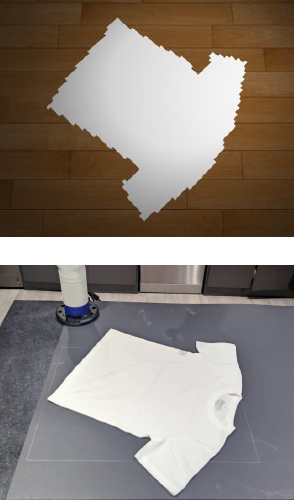} \\
    & \\
    (a) \small Rope & (b) \small Beans & (c) \small Cloth
    
\end{tabular}

\caption{Visualization. Bottom: deformable objects in the real world. Top: particle states rendered in simulation for different types of deformable objects.
}
\label{fig:particles_in_sim}
\end{figure}

\textbf{Visualizations} We provide visualizations of particles rendered in simulation based on real-world observations in \figref{fig:particles_in_sim}.

\textbf{Hyper-parameters} Table \ref{table:real_robot_hyperparameters} shows the hyperparameters used in our real robot experiments for rope pushing, bean sweeping and t-shirt folding.

\input{tables/real_robot_task_hyperparameters}

\textbf{Computation Time}
Table \ref{table:real_robot_computation_time} and Table \ref{table:real_robot_computation_time2} show the computation time for each method and for each task. For model parameters learning, each gradient update takes about 20 - 40 seconds depending on the demonstrated trajectory length.

\input{tables/robot_computation_time}

\textbf{Interaction between end-effector and particles}
The two gripper fingers are simulated as two boxes in the simulation. Collision between particles and boxes are well captured in the physics engine.

\subsection{Simulation Tasks}\label{append: simultion}

 \begin{figure}[ht]
\centering
	\begin{tabular}{c c}
    \includegraphics[width=0.395\linewidth]{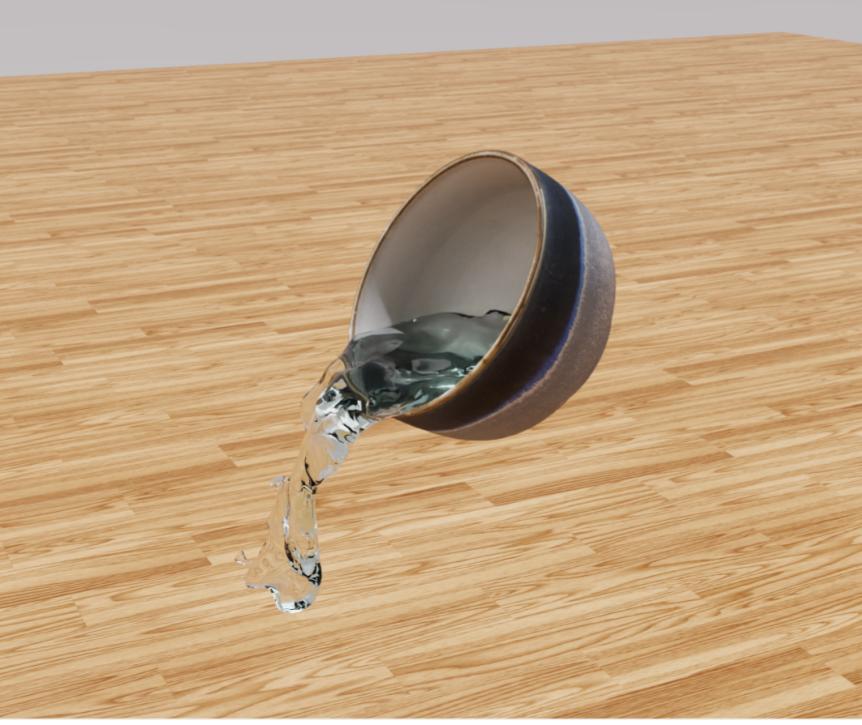} &

    \includegraphics[width=0.4\linewidth]{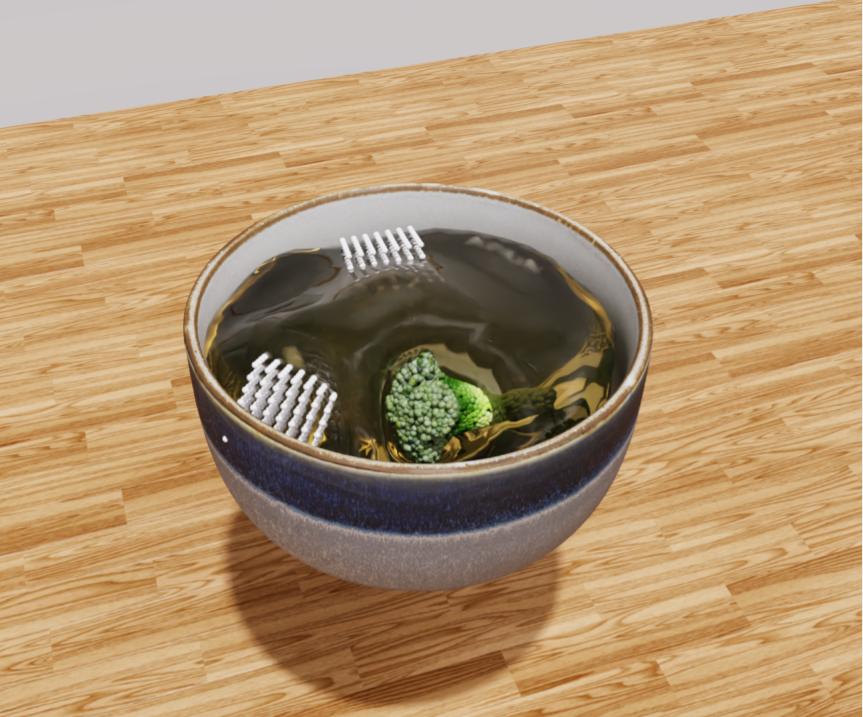}  \\
    \\
    (\subfig a) \small Pour water$^\dagger$ & (\subfig b) \small Pour Soup$^\dagger$ 
\end{tabular}
\caption{We illustrate different deformable object manipulation tasks in simulation. $\dagger$ We use the example images from DaxBench~\cite{Dax}.
}
\label{fig:tasks_in_sim}
\end{figure}

We present a comprehensive examination of simulations performed using the DaXBench platform~\cite{Dax}. The experiments encompass three distinct tasks, each with its own unique set of challenges and objectives.

\noindent\textbf{Pour Water} The Pour Water task requires the robot to pour a bowl of water into a target bowl, using control of velocity and rotation in Cartesian space. The robot holds the bowl and the action space is defined as a 6-dimensional vector, representing linear velocity along the $x$, $y$, and $z$ axes, as well as angular velocity along 3 axes. The task is designed to last for 100 steps.

\noindent\textbf{Pour Soup} The Pour Soup task is similar in concept to the Pour Water task, but with the added challenge of solid ingredients mixed in with the water. These ingredients create additional interactions between the liquid and solid components, potentially leading to spills. As with the Pour Water task, the action space is defined as a 6-dimensional vector, with the same linear and angular velocity controls. The task horizon is set at 120 steps.

Each of these tasks offers a unique set of challenges and objectives. Since Transporter is not deigned for low-level control tasks, we compare \anameDipac with four baselines in DaXBench, namely, ILD~\cite{chen2022imitation}, CEM-MPC, Diff-MPC, Diff-CEM-MPC.

ILD is the latest differentiable-physics-based IL method; it utilizes the simulation gradients to reduce the required number of expert
demonstrations and to improve performance.

Diff-MPC and Diff-CEM-MPC are two variants of Model Predictive Control that incorporate differentiable dynamics. Both methods optimize immediate rewards for each action using differentiable physics but differ in their initialization of action sequences. Diff-MPC starts with a random sequence, while Diff-CEM-MPC utilizes a sequence optimized by Cross-Entropy Method MPC (CEM-MPC).

\anameDipac uses a behavior cloning policy, with an image as input and action as output, as its initial policy. We use 1 GPU, 10 demonstrations for training initial policy, and search depths of the whole trajectory, 100, 120, and 70 for the 3 tasks. We outperformed all the baselines, demonstrating the effectiveness of our method.

%% file: tables/real_robot_task_hyperparameters.tex
\begin{table}[ht]
\centering
\caption{\small Hyperparameters for \anameDipac on the real robot tasks}
\fontsize{7}{7}\selectfont

\begin{tabular}{lccccc}
\toprule
Task &  $k$ + 1  & \# GPUs & \# Demos  & Search Depth & \# Particles     \\ 
\midrule

Rope Pushing  &  15 + 1  &  15  &   10  & 3  & 2000   \\
Bean Sweeping &  149 + 1   & 15   &  10 &  1  & 300  \\
Cloth Hanging  &  15 + 1  &  4  & 10 & 3  & 1000  \\
\bottomrule
\end{tabular}
\label{table:real_robot_hyperparameters}
\end{table}

%% file: tables/robot_computation_time.tex
\begin{table}[h]
\centering
\caption{\small Computation time on the real robot tasks}

\begin{tabular}{lccc}
\toprule
Method &  Rope Pushing  & Bean Sweeping & Cloth Hanging     \\ 
\midrule

Transporter  &  2s  &  2s  &   NA    \\
Diffusion Policy &  1s   & 1s   &  1s     \\
CEM-MPC  &  90s  &  90s  & 60s    \\
Dipac &  10s  &  10s  & 4s \\
\bottomrule
\end{tabular}
\label{table:real_robot_computation_time}
\end{table}

\begin{table}[h]
\centering
\caption{\small Detailed Computation Time}

\begin{tabular}{lccc}
\toprule
Method &  Rope Pushing  & Bean Sweeping & Cloth Hanging     \\ 
\midrule

Step Forward  &  0.05s  &  0.05s  &   0.03s   \\
Step Backward  &  0.2s   & 0.2s   &  0.1s     \\
Traj Optimization  &  0.6s  &  0.4s  & 0.3s    \\
\bottomrule
\end{tabular}
\label{table:real_robot_computation_time2}
\end{table}